\documentclass[pdflatex,sn-mathphys-num,noorcidlogo,smallextended,referee,lineno=false,nonlinenum]{sn-jnl}

\usepackage{tabularx}
\usepackage{float} 
\usepackage{graphicx}%
\usepackage{multirow}%
\usepackage{makecell}
\usepackage{amsmath,amssymb,amsfonts}%
\usepackage{amsthm}%
\usepackage{array}
\usepackage{mathrsfs}%
\usepackage[title]{appendix}%
\usepackage{textcomp}%
\usepackage{manyfoot}%
\usepackage{booktabs}%
\usepackage{algorithm}%
\usepackage{algorithmicx}%
\usepackage{algpseudocode}%
\usepackage{listings}%
\usepackage{geometry}  
\usepackage{minitoc}
\usepackage{titletoc}
\usepackage{fancyhdr}
\usepackage[table,xcdraw]{xcolor}
\usepackage{tikz,lipsum,lmodern}
\usepackage[most]{tcolorbox}
\usepackage{pdfpages}
\usepackage{hyperref} 
\usepackage{gensymb}
\tcbuselibrary{listings}
\usepackage{listingsutf8}
\usepackage{bibunits}
\usepackage{chngcntr}
\usepackage{hyperref}
\usepackage{cleveref}

\theoremstyle{thmstyleone}%
\theoremstyle{thmstyletwo}%

\theoremstyle{thmstylethree}%

\newcommand{\name}[1]{Earth-o1}
\newcommand{\olsname}[1]{LatentOS}

\begin{document}

\setlength{\bibsep}{0.5em}

\Crefname{figure}{Fig.}{Figs.}


\title[Earth-o1]{Earth-o1: A Grid-free Observation-native Atmospheric World Model}


\author[1,3]{\fnm{Junchao} \sur{Gong}}
\equalcont{These authors contributed equally to this work.}
\author[1,5]{\fnm{Kaiyi} \sur{Xu}}
\equalcont{These authors contributed equally to this work.}
\author[1,6]{\fnm{Wangxu} \sur{Wei}}
\equalcont{These authors contributed equally to this work.}
\author[1,7]{\fnm{Siwei} \sur{Tu}}
\equalcont{These authors contributed equally to this work.}
\author[1,7]{\fnm{Jingyi} \sur{Xu}}
\equalcont{These authors contributed equally to this work.}
\author[1]{\fnm{Zili} \sur{Liu}}
\author[8]{\fnm{Hang} \sur{Fan}}
\author[1]{\fnm{Zhiwang} \sur{Zhou}}
\author[1]{\fnm{Tao} \sur{Han}}
\author[1]{\fnm{Yi} \sur{Xiao}}
\author[1]{\fnm{Xinyu} \sur{Gu}}
\author[10]{\fnm{Zhangrui} \sur{Li}}
\author[1]{\fnm{Wenlong} \sur{Zhang}}
\author[1]{\fnm{Hao} \sur{Chen}}
\author[3]{\fnm{Xiaokang} \sur{Yang}}
\author[9]{\fnm{Yaqiang} \sur{Wang}}
\author[6]{\fnm{Lijing} \sur{Cheng}}
\author[8]{\fnm{Pierre} \sur{Gentine}}
\author[2]{\fnm{Wanli} \sur{Ouyang}}
\author[4]{\fnm{Feng} \sur{Zhang}}
\author[10]{\fnm{Zhe-Min} \sur{Tan}}
\author[1]{\fnm{Bowen} \sur{Zhou}}
\author*[1]{\fnm{Fenghua} \sur{Ling}}\email{lingfenghua@pjlab.org.cn}
\author*[2]{\fnm{Ben} \sur{Fei}}\email{benfei@cuhk.edu.hk}
\author*[1]{\fnm{Lei} \sur{Bai}}\email{baishanshi@gmail.com}

\affil*[1]{\orgname{Shanghai Artificial Intelligence Laboratory}, \orgaddress{\city{Shanghai}, \country{China}}}
\affil*[2]{\orgdiv{Department of Information Engineering}, \orgname{The Chinese University of Hong Kong}, \orgaddress{\city{Hong Kong}, \country{China}}}
\affil[3]{\orgdiv{School of Electronic Information and Electrical Engineering}, \orgname{Shanghai Jiao Tong University}, \orgaddress{\city{Shanghai}, \country{China}}}
\affil[4]{\orgdiv{Department of Atmospheric and Oceanic Sciences}, \orgname{Fudan University}, \orgaddress{\city{Shanghai}, \country{China}}}
\affil[5]{\orgdiv{School of Information Science and Technology}, \orgname{University of Science and Technology of China}, \orgaddress{\city{Anhui}, \country{China}}}
\affil[6]{\orgdiv{State Key Laboratory of Earth System Numerical Modeling and Application, Institute of Atmospheric Physics}, \orgname{Chinese Academy of Sciences}, \orgaddress{\city{Beijing}, \country{China}}}
\affil[7]{\orgdiv{College of Computer Science and Artificial Intelligence}, \orgname{Fudan University}, \orgaddress{\city{Shanghai}, \country{China}}}
\affil[8]{\orgdiv{Department of Earth and Environmental Engineering}, \orgname{Columbia University}, \orgaddress{\city{New York}, \state{NY}, \country{USA}}}
\affil[9]{\orgname{Chinese Academy of Meteorological Sciences}, \orgaddress{\city{Beijing}, \country{China}}}
\affil[10]{\orgdiv{School of Atmospheric Sciences}, \orgname{Nanjing University}, \orgaddress{\city{Nanjing}, \country{China}}}


\abstract{
Despite the unprecedented volume of multimodal data provided by modern Earth observation systems, our ability to model atmospheric dynamics remains constrained. Traditional modeling frameworks force heterogeneous measurements into predefined spatial grids, inherently limiting the full exploitation of raw sensor data and creating severe computational bottlenecks. Here we present Earth-o1, an observation-native atmospheric world model that overcomes these structural limitations. 
Rather than relying on conventional atmospheric dynamical modeling systems or traditional data assimilation, Earth-o1 directly learns the continuous, three-dimensional physical evolution of the Earth system from ungridded observational data. 
By integrating diverse sensor inputs into a unified, grid-free dynamical field, the model autonomously advances the atmospheric state in space and time. We show that this fundamentally distinct paradigm enables direct, real-time forecasting and cross-sensor inference without the overhead of explicit numerical solvers. In hindcast evaluations, Earth-o1 achieves surface forecast skill comparable to the operational Integrated Forecasting System (IFS). These results establish that continuous, observation-driven world models---a new class of fully observation-native geophysical simulators---can match the fidelity of established physical frameworks, providing a scalable data-driven foundation for a digital twin of the Earth.
}

\maketitle

\section{Main}
\label{sec:main}

Observations are the primary means by which the Earth system can be directly measured and explored, and they underpin a major part of our understanding in the atmospheric and climate sciences~\cite{allen2025end,alexe2024graphdop,feng2026can}. Advances in satellite instruments, radar systems, and ground-based networks have produced a rapidly expanding volume of multi-source Earth observations~\cite{yang2017introducing,bessho2016introduction,abreu2012description}. These data support long-term climate monitoring, extreme-event prediction, and real-time early-warning systems~\cite{zhang2023skilful,ravuri2021skilful}. 
Notwithstanding these advances, the current use of observations remains constrained. Most applications depend on model grid structure, where raw data must be projected onto predefined spatial grids and variables~\cite{bi2023accurate,lam2023learning,chen2023fengwu,han2024fengwu}. 
In this process, more than 90\% of the available satellite information is thinned or discarded~\cite{mcnally2024data}. 
This rigid framework not only strips away fine-scale signals and cross-sphere interactions, but also filters out valid observations that are incompatible with predefined grid structures~\cite{rabier2000ecmwf,hersbach2020era5}. Such aggressive data curation is primarily enforced to reduce computational overhead and stay within a practical budget.
Consequently, current grid-based systems preserve physical consistency and forecast skill but at the cost of discarding observational heterogeneity and complexity essential for improving Earth-system understanding, and potentially prediction.
These limitations have led to a growing interest in data-driven approaches for representing atmospheric dynamics.
Recent AI-based forecasting systems, such as FengWu~\cite{chen2023fengwu,ling2024fengwu,han2024fengwu,guo2025data}, PanGu-Weather~\cite{bi2023accurate}, and GraphCast~\cite{bi2023accurate}, have demonstrated that neural networks can reproduce large-scale flow evolution with remarkable skill. Yet these models operate entirely within the pre-defined spatial grids in numerical model products. By training on reanalysis fields, they inherit the foundational biases of predefined spatial grids and variables, bypassing most of the information contained in raw observations. 
Attempts to improve the use of observations include FengWu-4DVar~\cite{xiao2023fengwu}, Aadvark Weather~\cite{allen2025end}, and et~al.~\cite{chen2023towards,chen2024fnp}, which accelerates the mapping of multi-source measurements into model space through neural transformations~\cite{allen2025end}. This reduces latency in data assimilation but is also constrained by the same pre-defined spatial grids as in numerical model.
More observation-oriented developments, such as GraphDOP, move further by constructing latent embeddings directly from heterogeneous observations~\cite{alexe2024graphdop}. This offers an important step toward observation-driven atmospheric learning, but it is still constrained by fixed grid structures in the data itself, making it difficult to capture localized phenomena in complex terrain, station-specific signals, and extreme events. 
Here we present \textbf{\name}, an observation-native atmospheric world model capable of resolving these inherent structural constraints, enabling the reconstruction and prediction of arbitrary 3D positions on Earth and revealing the real Earth even in areas where traditional grids are ineffective.
As a world model, it consists of three modules: reconstruction, prediction, and inversion~\cite{agarwal2025cosmos}. In the reconstruction module, \name~ constructs a unified \textbf{Latent Observation Space (\olsname~)} directly from multi-source heterogeneous observational data. Instead of projecting observations onto fixed grids, \name~ learns the continuous, three-dimensional physical evolution of the Earth system directly from ungridded observational data, forming a continuous, high-dimensional latent manifold. This observation-driven representation avoids explicit physical parameterizations, preserves the intrinsic structure of the observational signal, and enables inference at flexible spatial resolutions and arbitrary 3D positions. Based on this unified latent manifold, the prediction module can learn the spatio-temporal evolution of the atmospheric state from observations without relying on external dynamic cores. The inversion module establishes the inherent correlations among world observations and transfers real-time observations to downstream high-value products such as sea ice and pollutants.
Our results show that the world model paradigm represented by \name~ enables three core capabilities: First, high-precision reconstruction of atmospheric states at arbitrary points, the resulting \olsname~ extends the regional observation signals to the global, making signals of GEO signals cover the whole Earth, achieving the reconstruction of physical quantities at arbitrary points on Earth with the ability to capture extreme events surpassing that of ERA5. Second, real-time forecasting of atmospheric sensor states, as \name~ directly takes observations as input instead of analysis data, its prediction is real-time. Further, \name~ captures essential atmospheric dynamics, achieving forecast skill comparable to the operational IFS model~\cite{wedi2015modelling} with the capture of extreme event transitions comparable to or even exceeding that of the IFS model.
Third, high-value cross-modal inference, the integrative nature of the \olsname~ supports direct retrieval of cross-component variables from diverse data streams, including sea-ice evolution to atmospheric composition monitoring.
In summary, these outcomes confirm that continuous, observation-driven world models match the fidelity of established physical frameworks, offering a scalable data-driven foundation to advance the development of an Earth digital twin.

\section{Results}
\subsection*{Overview of Earth-o1}

\begin{figure}[ht]
    \centering
    \includegraphics[width=1.0\linewidth]{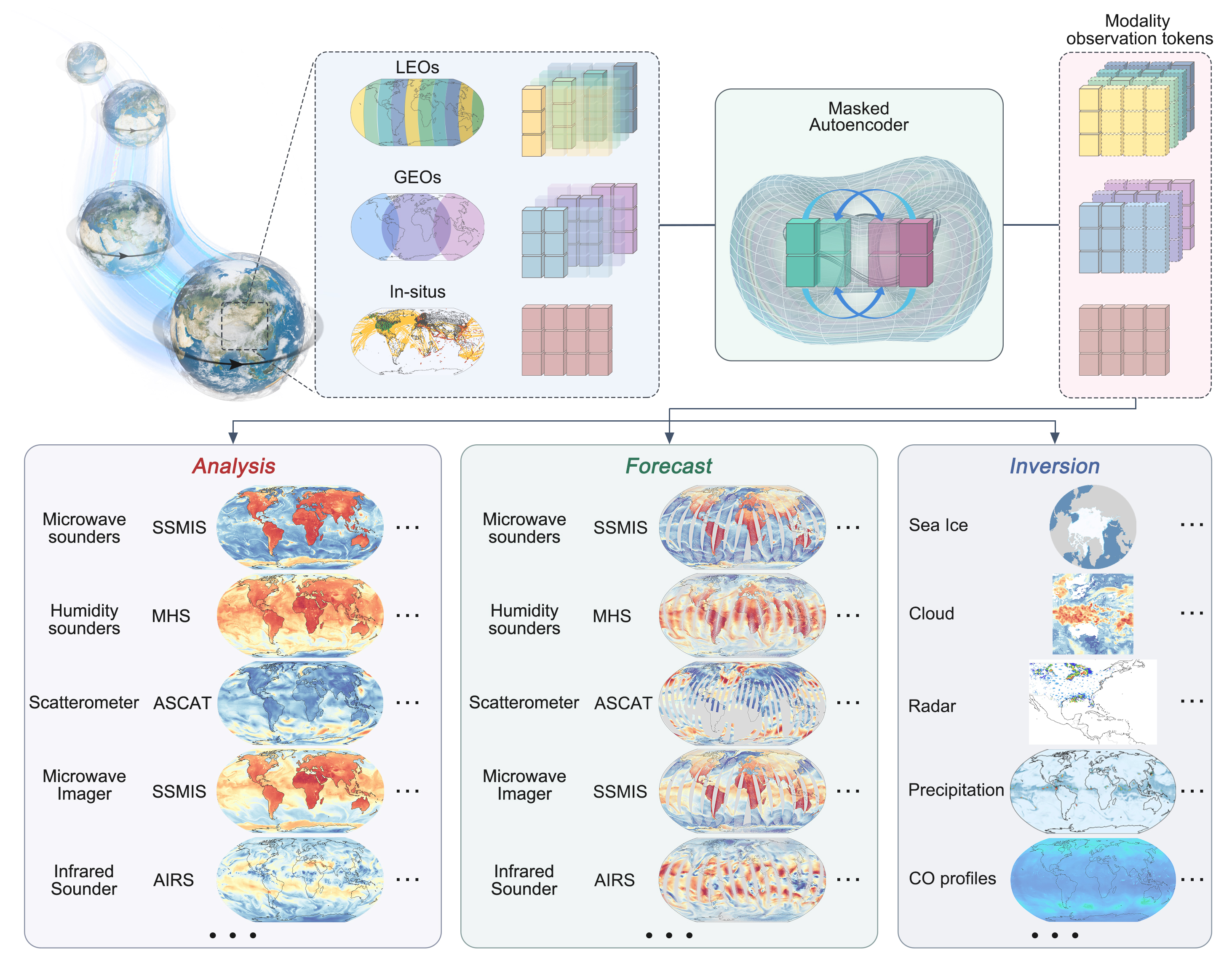}
    \vspace{-0.7cm}
    \caption{Overview of \name~. After tokenization, multi-modal heterogeneous data from LEOs, GEOs and in-situ platforms are trained via a Masked Autoencoder, where representations are mutually integrated to generate modality observation tokens. \textbf{Analysis}: these latent observation tokens support multi-sensor analysis tasks, including microwave and infrared remote sensing. \textbf{Forecast}: facilitate the forecasting of observation signals. \textbf{Inversion}: enable the inversion of downstream geophysical products; as illustrated in the figure, high-value products such as sea ice concentration and CO concentration are inverted.}
    \label{fig1}
\end{figure}

To operationalize a fully observation-driven paradigm, \name~ integrates diverse global observing systems into a unified representation of the atmospheric state (\Cref{fig1}).
Rather than projecting dense satellite measurements and sparse in-situ observations onto a fixed analysis grid, the framework introduces a shared \olsname~ that embeds heterogeneous observational information into the latent observation space. 

This space is learned through multimodal masked autoencoding~\cite{vandal2025global,mizrahi20234m}, enabling the joint integration of satellite observations spanning microwave, infrared, and visible bands with in-situ measurements across observing systems. Training is performed on a petabyte-scale corpus encompassing Level-1 records from ten polar-orbiting (LEO) instruments
, three geostationary (GEO) platforms
, and seven disparate in-situ networks
. During training, the model is tasked with reconstructing withheld observations across instruments and platforms, forcing the latent representation to remain robust to data gaps, uneven spatial-temporal coverage, and sensor-specific limitations while preserving physically consistent relationships among heterogeneous measurements. This fusion process constrains \olsname~ to function as a continuous, sensor-agnostic manifold that transcends the limitations of individual observing systems.

Within this unified LatentOS, the temporal evolution is learned using a Transformer-based sequence model~\cite{yang2024qwen2,bai2025qwen2,yang2025qwen3} that advances the atmospheric state forward in time. Rather than relying on prescribed numerical equations, the model infers dominant transport and interaction patterns from the observational record, enabling the representation of multi-scale atmospheric variability within a single evolving state. Since this latent evolution is independent of any particular model variable, the predicted state can be mapped to different physical quantities through lightweight, task-specific decoding modules. In this way, \olsname~ functions as a common intermediate state supporting multiple observation operators, allowing diverse Earth system products—such as precipitation, cloud properties, or surface conditions—to be derived from a shared, dynamically consistent representation.

\begin{figure}[htbp]
    \centering
    \includegraphics[width=1.0\linewidth]{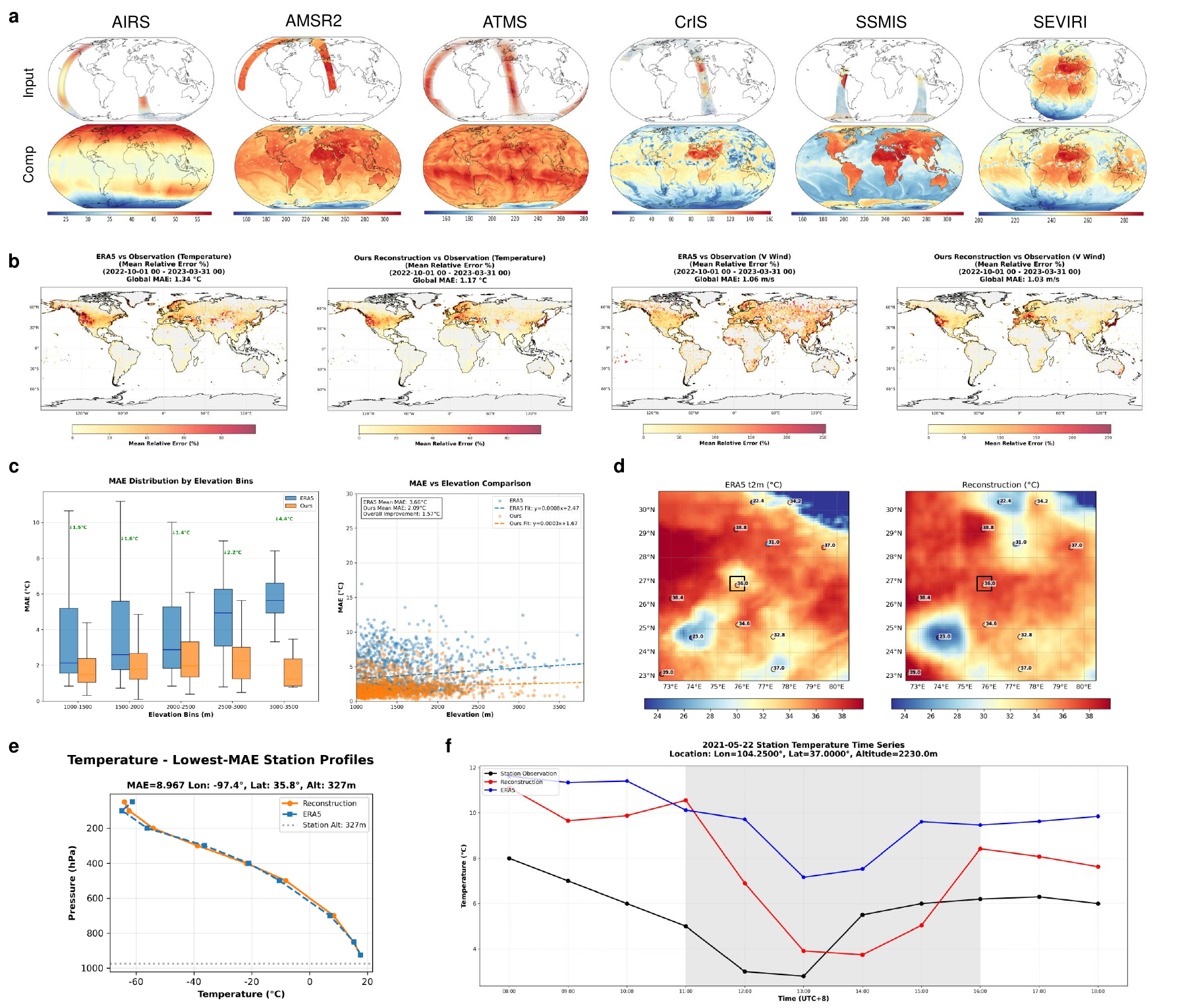}
    \vspace{-0.7cm}
    \caption{
    \textbf{a,} Illustrations of the raw data and completion maps from six of the satellite instruments. The unobserved areas in the swath instruments have been filled in latent observation space, yielding a continuous global-scale observational field for all types of observation data. 
    \textbf{b,} Error comparison of \name~ and ERA5 at global surface stations for the V wind and temperature variables. Reconstruction results from \olsname~ achieve higher accuracy.
    \textbf{c,} Analysis of reconstruction error variations with elevation. For temperature, we compare the variations of our reconstruction, ERA5, and in-situ surface observations with elevation, including error distributions across elevation bins, regression functions of mean error against elevation as a function of elevation. 
    \textbf{d,} Visualization of continuous field reconstruction from sparse observations, using the Rajasthan Heatwave during 1–11 June 2021 as a case study. 
    The continuous surface temperature fields are reconstructed by \name~ from \olsname~ and ERA5 t2m over the region at 12:00 UTC on 4 June 2021. 
    As highlighted by the black box, \name~ yields smaller temperature errors near real observations, demonstrating the capability of our method in reconstructing continuous fields from sparse observations. 
    \textbf{e,} Temperature profiles at an observation site. We compare profiles above the station at 97.4$^\circ$W, 35.8$^\circ$N at 00:00 UTC on 25 May 2021. From 1000 hPa to 50 hPa, the temperature profile trend is consistent with ERA5, illustrating the potential of Earth-o1 in modeling vertical atmospheric variations.
    \textbf{f,} 
    Taking the extreme temperature change event of the 2021 Baiyin Marathon as an example, we plot the variation curves of our reconstruction results, ERA5, and observed temperature~\cite{zhang2021lessons} within 10 hours. The reconstruction results of \name~ exhibit a closer temporal dynamic consistency with the observations, especially during the extreme cooling period from 11:00 to 16:00.
    }
    \label{fig2}
\end{figure}

\subsection*{Earth-o1 reconstructs spatially complete global observations}
The primary capability of \name~ is its ability to integrate diverse observations into a physically coherent global atmospheric and cross-sphere state. As illustrated in Fig.~\Cref{fig2}a, the system jointly assimilates heterogeneous Earth Observation (EO) measurements from both polar-orbiting and geostationary platforms, producing a globally complete reconstruction despite the discontinuity of the observation. The resulting fields exhibit spatial structures that are physically plausible at synoptic scales, indicating that the reconstruction is not achieved through simple spatial interpolation, but through the effective fusion of complementary information provided by different instruments. This interpretation is further supported by sampling experiments (in Supplementary). When the available observations are reduced to 20\% or even 10\% of the reference sampling density, the reconstruction accuracy remains largely preserved. Significant structural degradation emerges only when the sampling density falls below 5\%. Together, these results indicate that \name~ captures the low-dimensional manifold governing atmospheric observational variability, enabling robust recovery of the global state from sparse and irregular observations.

Beyond satellite radiances, the latent representation learned by \name~ consistently integrates in-situ surface measurements within the same global state. Ground stations are treated as sparse but high-accuracy constraints embedded in the continuous atmospheric field, enabling direct coupling between kilometre-scale satellite observations and point-scale surface conditions.
As shown in \Cref{fig2}b, the reconstructed near-surface air temperature at station locations achieves a mean absolute error of 1.17 K, and 1.03 m/s for V wind speed, matching or exceeding the performance of the ERA5 reanalysis.
Further analysis of the reconstruction reveals that the advantages stem from the higher robustness of our method to variations in surface elevation. As shown in \Cref{fig2}c, although the reconstruction variance increases with elevation, the modeling capability of \name~ for surface air temperature is less affected by elevation compared with ERA5, with a coefficient of only 0.0003 for the growth of MAE with elevation.
\name~ also shows promising potential in characterizing atmospheric profiles when vertical variations are extended to larger-scale pressure levels. As illustrated in \Cref{fig2}e, \name~ yields similar temperature profiles to ERA5 above the same station between 1000 hPa and 50 hPa.
Beyond its ability to accurately incorporate point-scale constraints in space, 
\name~ achieves performance comparable to ERA5 in capturing temporal dynamics.
During a sharp temperature drop event at the Baiyin Marathon~\cite{zhang2021lessons}(\Cref{fig2}f), 
our results track the air temperature more accurately than even ERA5 over the 10-hour period, with the lowest temperature difference below 4$^\circ\text{C}$.
This cross-scale consistency allows fine-scale thermal structures to emerge without explicit downscaling or surface-specific parameterization. By querying the reconstructed state at arbitrary spatial locations, \name~ resolves localized temperature gradients associated with heterogeneous land surfaces and better retrieves localized extreme values. 
As shown in \Cref{fig2}d, the continuous field inferred by \name~ is tightly coupled with the observed points, and the temperature gradients are effectively propagated to the surrounding regions. Taking the black box as an example, the temperature reconstructed by \name~ within the box is close to the observed value of 36$^\circ$C, while ERA5 only yields approximately 30$^\circ$C. 
These results demonstrate the superior capability of \name~ in fusing observations across instruments, and spatiotemporal scales.

\subsection*{Learning atmospheric dynamics within Earth-o1}

Having shown that \name~ reconstructs a physically coherent atmospheric state from diverse observations, we next examine whether this reconstructed state also preserves the temporal evolution of atmospheric variability. If the \olsname~ provides a meaningful description of the atmosphere, its evolution should exhibit short-range predictability consistent with observed dynamical behaviour.

\begin{figure}[htbp]
    \centering
    \includegraphics[width=1.0\linewidth]{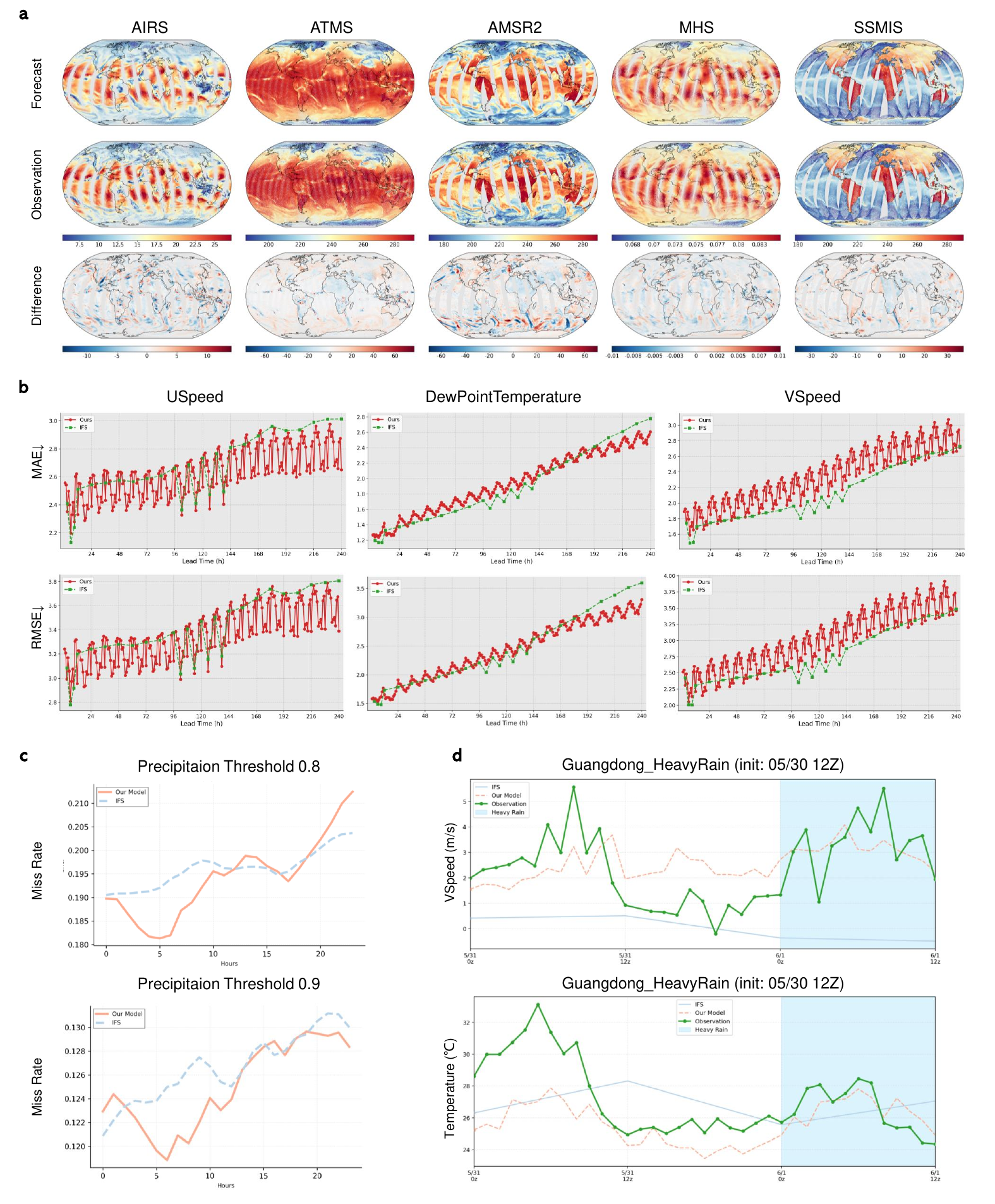}
    \vspace{-0.7cm}
    \caption{
    \textbf{a,} Global 0 to 12-hour lead-time forecasts of \name~, associated observation fields(from 06:00 to 18:00 UTC 29 on November 2022), and forecast minus observation differences for representative LEO satellite channels (ARIS channel 10, ATMS channel 4, AMSR2 channel 10, MHS channel 4, SSMIS channel 1).
    \textbf{b,} Forecast errors of global station observations, presenting results for three surface variables, including speed of U wind, dew point temperature, and speed of V wind.
    \textbf{c,} Evaluation of 24-hour precipitation forecast capability. Using the ATMS precipitation product as the truth value, we evaluated the extreme precipitation results of \name~ and IFS 9 km forecasts(downsampled to 0.16$^\circ$) initiated every 12 hours between 00:00 UTC on 1 July 2023 and 00:00 UTC on 20 July 2023 at the 80\% and 90\% thresholds. The result is calculated in a smooth window of 5.
    \textbf{d,} Visualization of extreme event forecasts. Taking the heavy rain in Guangdong as an example, we evaluated the error curves between the predicted values of surface variables and the real observations within 36 hours.
    }
    \label{fig3}
\end{figure}

\Cref{fig3}a shows that 0-12 hours forecasts initiated from the \olsname~ reproduce the dominant large-scale circulation patterns across five satellite observation sources. The persistence and displacement of synoptic-scale features are consistent with the observed atmospheric evolution, indicating that the reconstructed state retains dynamically relevant information rather than representing a static spatial fit. This behaviour suggests that the \olsname~ captures not only instantaneous structure but also the evolution of the underlying flow. This dynamical consistency is further reflected in the quantitative forecast skill. 
For near-surface \textit{in-situ} variables (\Cref{fig3}b), forecasts based on \olsname~ achieve accuracy comparable to the IFS system. In terms of Mean Absolute Error (MAE), both \name~ and the IFS system yield identical MAE values within the first 18 forecast hours, with a minimum value of 2.2. For the wind speed (USpeed) during the lead time of 18–96 hours, \name~ delivers a slightly lower MAE of approximately 2.4, outperforming the IFS system at around 2.6. \name~ also exhibits superior performance for extended forecast horizons. When the lead time is extended to 192–240 hours, \name~ achieves an MAE of roughly 2.7 for USpeed and 2.5 for DewPointTemperature, further exceeding the performance of the IFS system.
This indicates that the observational constraints encoded in the \olsname~ remain effective and consistent as the system evolves forward in time.

Since LatentOS exhibits short-range predictability consistent with observed dynamical behaviour, the next key question is whether such predictability can effectively capture extreme weather. We demonstrate the capability of \name~ in forecasting extreme weather from two perspectives: global-scale statistical evaluation and local extreme weather events. \Cref{fig3}c presents the comparison of 24-hour precipitation forecasts between \name~ and 0.09° IFS predictions (both downsampled to 0.16$^\circ$), using ATMS precipitation products as ground truth, over the extreme distributions at the 80\% and 90\% thresholds. At the 90\% threshold, the Miss rate of our forecast is lower than that of IFS for the first 10 hours and matches IFS afterward. In addition to the global statistical analysis, we also investigate forecast performance during regional extreme weather events. As shown in \Cref{fig3}d, before and during the heavy rainfall event in Guangdong Province, V wind speed and temperature predicted by \name~ are closer to observations for near-surface variables.

\begin{figure}[htbp]
    \centering
    \includegraphics[width=1.0\linewidth]{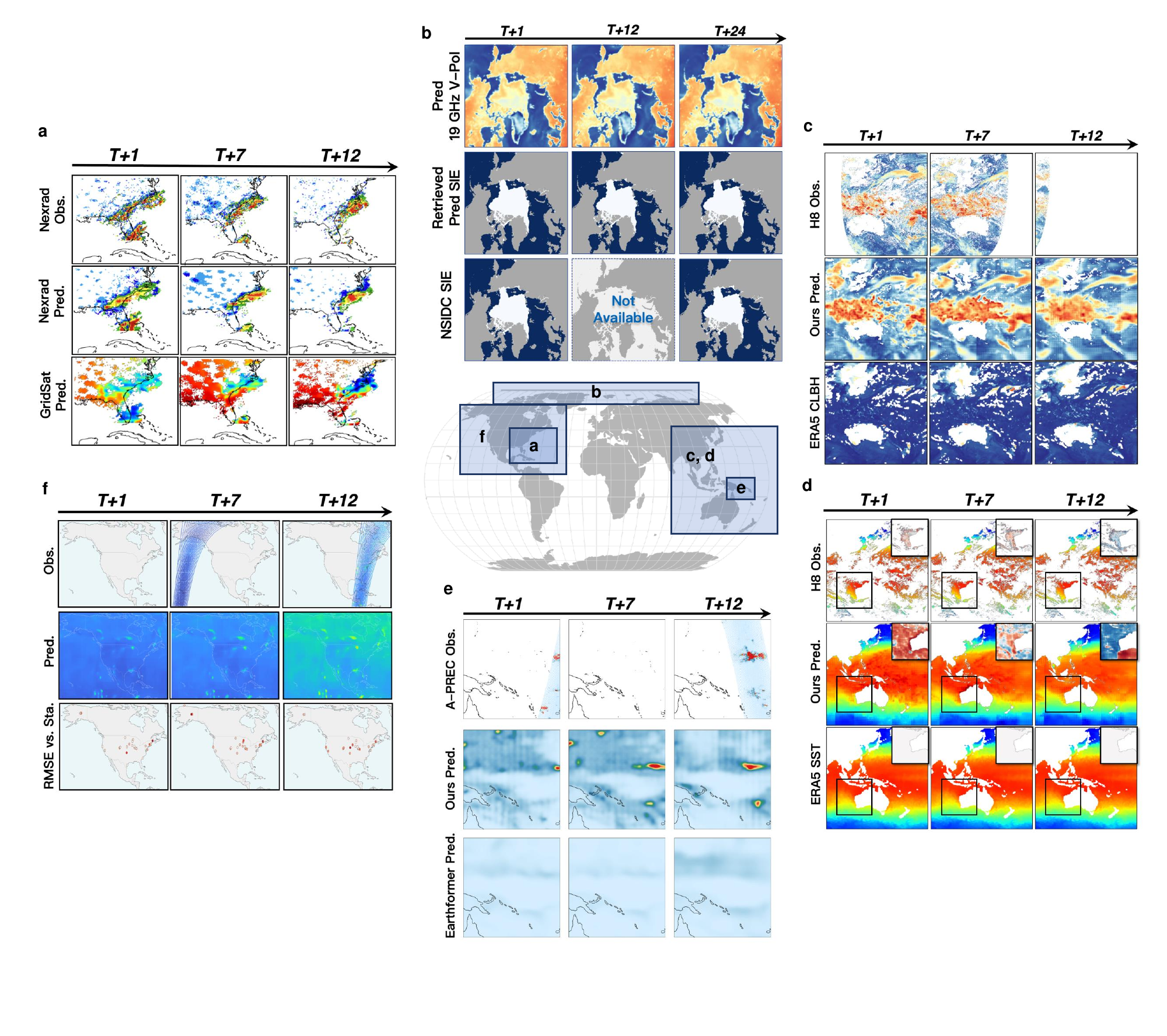}
    \vspace{-0.5cm}
    \caption{
    \textbf{a,} The radar composite reflectivity forecasts over North America derived from GRIDSAT inputs exhibit a high degree of consistency with the ground-truth radar observations (initialized forecasting time at 12:00 UTC on 7 June 2017). 
    \textbf{b,} Benefiting from the temporal resolution of satellite observations and the capability of observation completion, our method can retrieve hourly-scale variations in SIC, which is not available in reanalysis data, and shows high consistency with the NSIDC Sea Ice Extent (initialized forecasting time at 00:00 UTC on 15 October 2022 with brightness temperature observations from SSMIS). 
    \textbf{c,} Compared with the Himawari-8 cloud-top height (CLTH) product, our forecasts of CLTH over the East Asia–Pacific region, derived from Himawari-8 L1 inputs, exhibit greater spatial completeness and additionally provide cloud-top information—whereas ERA5 supplies only cloud-base information (initialized forecast time at 00:00 UTC on 1 April 2017).
    \textbf{d,} From the sea surface temperature (SST) forecasts over the East Asia–Pacific region, retrieved from Himawari-8 Level-1 forecast data, we also observe that our method reconstructs spatially complete, dense fields while faithfully capturing the rise-then-fall SST evolution consistent with the Himawari-8 GT product—a behavior that ERA5 fails to depict due to its insufficient temporal resolution (initialized forecast time at 00:00 UTC on 1 April 2017).
    \textbf{e,} We compare the precipitation inversion of our ATMS forecast with direct precipitation predictions from Earthformer, alongside the ATMS precipitation product. The example is presented over the region with longitude range $[140^\circ E, 179^\circ E]$ and latitude range $[20^\circ S, 20^\circ N]$, initialized at 12:00 UTC on 1 August 2022.
    \textbf{f,} We illustrate the CO inversion over North America from the joint AIRS--AMSU-A setting based on the learned fusion space, together with station measurements. By first completing the AIRS and AMSU-A observations and then performing multi-sensor joint inversion, \name~ reconstructs spatially continuous CO fields that better align with station observations than the AIRS-only setting, even though the two instruments are not co-orbital in the original observation space.
    }
    \label{fig4}
\end{figure}

\begin{figure}[htbp]
    \centering
    \includegraphics[width=1.0\linewidth]{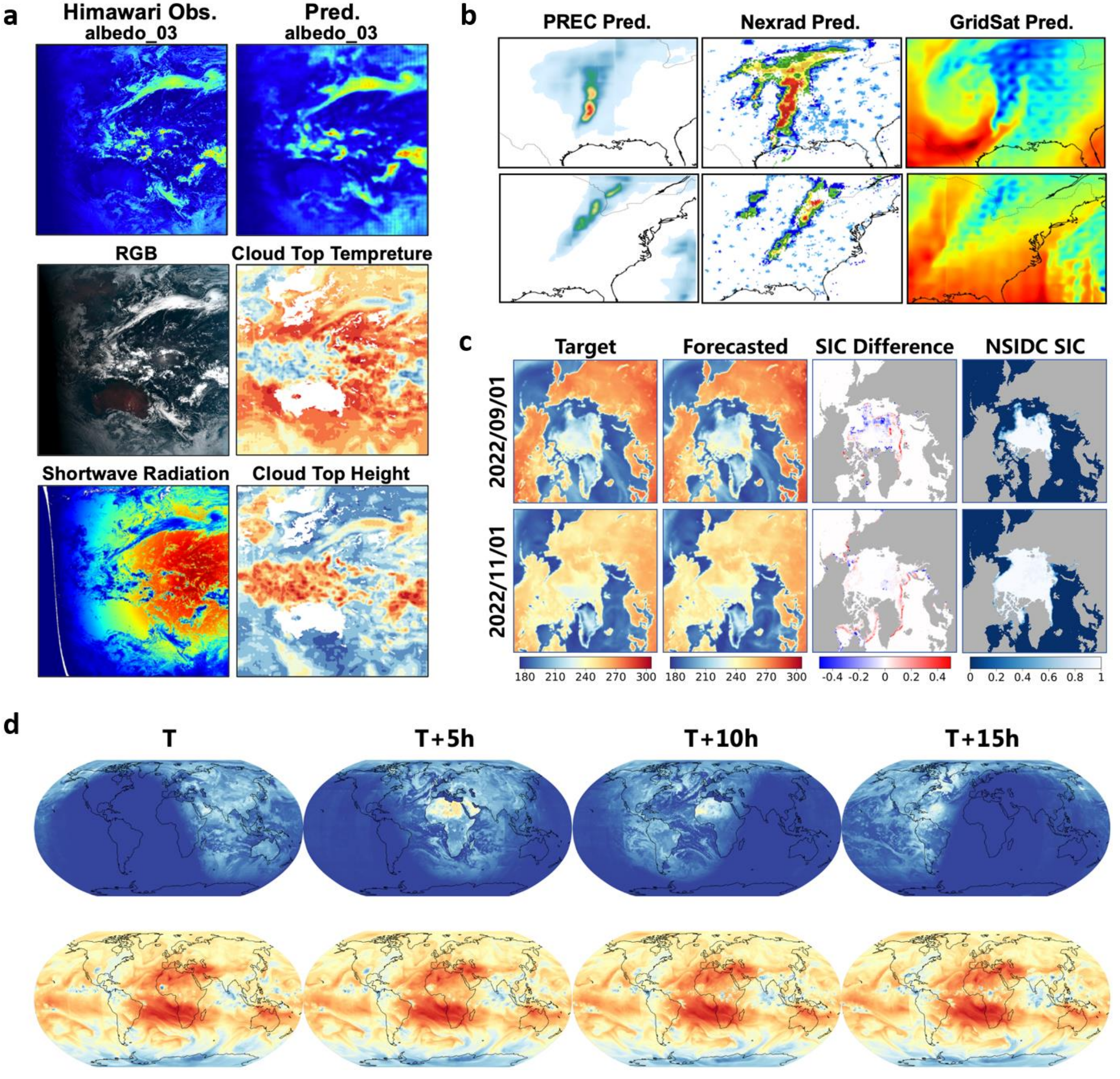}
    \vspace{-0.7cm}
    \caption{
    \textbf{a,} 
    The consistency of spatial patterns among forecasted reflectance at T+1, Himawari’s cloud products, and multi-band radiance.
    \textbf{b,} From the central storm structure predicted in the North American radar composite reflectivity, to the precipitation retrieved from ATMS, and the cloud-top patterns captured by the geostationary satellite, the three modalities exhibit remarkable spatial and structural consistency. 
    \textbf{c,} When predicting and retrieving SIC solely from SSMIS brightness temperatures, the predicted brightness temperature distributions closely resemble the target satellite observation, and the overestimation errors in retrieved SIC exhibit similar spatial patterns, concentrated predominantly along the sea-ice edge (initialized forecasting time at 00:00 UTC on 1 September 2022 and 00:00 UTC on 1 November 2022).
    \textbf{d,} After being fused with the latent observation space, observations across 16 hours from the geostationary satellite SEVIRI, are shown: the first row represents the shortwave infrared channel at 1.6 $\mu m$, and the second row represents the infrared band at 6.2 $\mu m$.
    }
    \label{fig5}
\end{figure}

\subsection*{Derivation of multi-sphere earth variables from the Latent Observation Space}

Atmospheric observations are central to a wide range of applications in meteorological and climate sciences. Through a unified latent observation space, \name~ provides a consistent observational foundation that supports these applications. For example, the multifrequency microwave radiances observed by the polar-orbiting satellite instrument SSMIS are largely insensitive to cloud cover and polar night conditions, enabling robust investigations of polar sea-ice variability based on the stable and distinct microwave emissivity contrast between sea ice and open water~\cite{beitsch2014comparison}. Similarly, thermal infrared brightness temperatures from the geostationary Himawari satellite can be used to characterize oceanic thermal radiation and to derive sea surface temperatures that closely approximate in situ conditions~\cite{kurihara2016sea}. 

To evaluate the generality of \name~ in supporting downstream scientific tasks, we compiled a diverse set of core application products, including precipitation estimates from ATMS~\cite{surussavadee2010npoess}, atmospheric composition retrievals from AIRS~\cite{smith2020climcaps}, and cloud-top height products from Himawari~\cite{ri2022cloud}, together with multi-model analysis datasets spanning atmospheric and sea-ice systems. This comprehensive collection enables a systematic assessment of \name~ across heterogeneous observational targets and application domains.

The advantages of \name~ manifest in three key aspects: extending the intrinsic spatiotemporal scales of raw observations, bridging joint information across diverse observing systems, and compensating for the limitations of existing analysis datasets. In terms of spatiotemporal coverage, \name~ extends ATMS-based precipitation products from multi-day revisit cycles to hourly monitoring, while expanding polar-orbiting observations to near-global coverage (\Cref{fig4}e). In addition, the completed and forecasted brightness temperature fields generated by \name~ enable the retrieval of continuous pan-Arctic sea-ice distributions (\Cref{fig4}b). These gains in spatiotemporal resolution establish a common foundation for linking disparate observing platforms. This capability is further reflected in atmospheric composition estimation. Although AIRS and AMSU operate on non-coincident orbits, their observations can be jointly exploited within the latent observation space to estimate CO (\Cref{fig4}f), O$_3$, CH$_4$, and OLR (in Supplementary), yielding performance improvements of over 30$\%$ in 5-day forecasts compared with single-instrument prediction. 
Moreover, as a key advantage of the foundation model, the unified latent space facilitates cross-instrument information sharing. This enables the reconstruction of active radar reflectivity signals solely from passive geostationary satellite observations (\Cref{fig4}a).
Beyond individual instruments, these properties allow \name~ to complement the temporal granularity and information content of conventional analysis products. Sea-ice extent from NSIDC SIE (\Cref{fig4}b) and sea surface temperature from ERA5 are refined from daily to hourly resolution, with reconstructed sea surface temperature variations closely matching observed values (\Cref{fig4}d). 
Owing to the unified latent observation space, \name~ also generates products that are absent from reanalysis datasets, such as cloud-top properties (\Cref{fig4}c), whereas ERA5 provides only cloud-base products.

\subsection*{Multi-perspective consistency inherit in \name}
Self-consistency represents another fundamental property of the latent observation space. This property spans across spectral, instrumental, modal, and spatiotemporal dimensions. For example, cloud patterns derived from HIMAWARI L1 albedo03, RGB composites~\cite{bessho2016introduction,shimizu2020introduction}, and radiative products are spatially coherent with the predictions of \name, indicating that \olsname~ preserves stable physical structures across multi-spectral representations within a single instrument (\Cref{fig5}a). This consistency extends to multi-instrument and cross-modal settings. Predicted centres of intense precipitation are spatially collocated with radar echo maxima and satellite-observed cold cloud-top regions, demonstrating that dynamically coherent features are aligned in \olsname~ despite originating from distinct observing modalities (\Cref{fig5}b). Similarly, sea-ice concentration (SIC) and extent (SIE) independently retrieved from AMSR-2 and SSMIS polarisation channels show highly consistent spatial distributions, with discrepancies primarily confined to marginal ice zones. Corresponding brightness temperature forecasts exhibit comparable physically realistic patterns, further supporting inter-instrument consistency within \olsname~ at similar frequency bands from multi-instruments (\Cref{fig5}c). Beyond instrumental and modal dimensions, \olsname~ also maintains continuity across regions and time. Shortwave infrared observations from the geostationary SEVIRI instrument display periodic spatial shifts associated with Earth’s rotation, yet remain temporally self-consistent in \olsname~ (\Cref{fig5}d). In addition, atmospheric structures captured by the SEVIRI water vapour channel are continuously represented across the transition between South and North America, highlighting robust cross-regional coherence (\Cref{fig5}d). Together, these results demonstrate that \olsname~ sustains physically consistent representations across heterogeneous and diverse observations.

\subsection*{Scaling of Earth-o1}
\begin{figure}[ht]
    \centering
    \includegraphics[width=\linewidth]{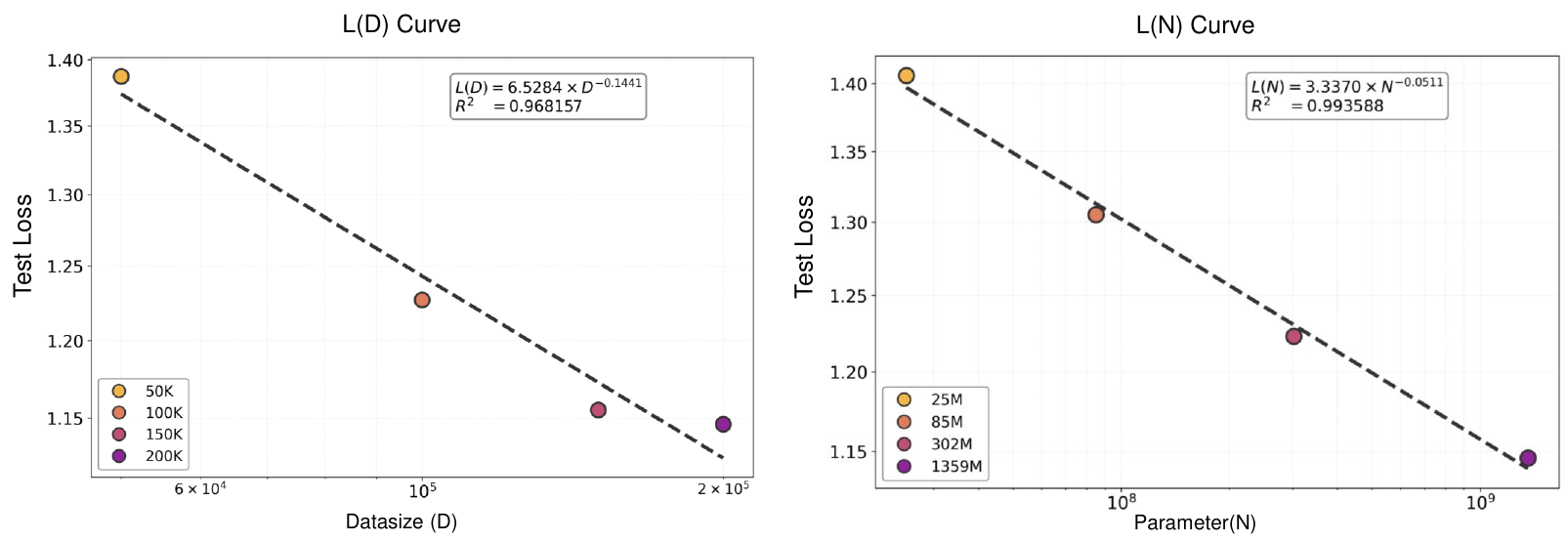}
    \caption{Scaling law of dataset size and parameters for satellite fusion.}
    \label{fig:scaling-loss}
\end{figure}

We conducted a series of scaling experiments~\cite{kaplan2020scaling,hoffmann2022training} on the observation fusion module, to verify its scaling capability with respect to data volume and model parameter count. For the experimental setup: first, we fixed a 1.3B-parameter model and swept across four dataset sizes (50k, 100k, 150k, and 200k), with each training run terminated at the onset of overfitting to characterize the scaling trend of loss against dataset size. Second, to derive the dependence of model performance on parameter count, we swept across four model scales (25M, 85M, 302M, and 1359M parameters) using a sufficiently large dataset, with all experimental groups fixed at 200k training steps. The  observation fusion module exhibits favorable scalable properties (\Cref{fig:scaling-loss}):
\begin{align}
    &L(D)=6.5284 \times D^{-0.1441}, \\
    &L(N)= 3.3370 \times N^{-0.0511}.
\end{align}
As model scale expands and training data volume increases, the reconstruction loss of \name~ decreases continuously. \name~ holds the potential to further enhance its capability in modeling the Earth system by leveraging multi-source observational data and larger model scales in the future.

\section{Discussion}
\name~ introduces a novel latent observation space paradigm to develop a purely observation-driven Earth modeling system.
By bypassing the traditional reliance on rigid grid-based reanalysis data, \name~ preserves the high-fidelity information inherent in multisource raw observations to drive monitoring, understanding, and forecasting of diverse tasks.
By integrating over 20 types of heterogeneous observations, \name~ overcomes the spatiotemporal limitations of any single sensor to enable real-time, high-frequency global monitoring of the Earth system. 
Through the fusion of atmospheric, oceanic, and polar data, the latent observation space supports tasks such as atmospheric composition forecasting, sea surface temperature prediction, and polar sea ice forecasting.

Improvements are possible along several axes. First, \name~ can be extended to incorporate uncertainty, for example, by constructing a latent observation space that includes uncertainty, using ensemble forecasting for predictions, and applying probabilistic modeling to downstream tasks~\cite{price2025probabilistic,ravuri2021skilful}. Second, expanding the latent observation space to integrate more instruments and modalities—such as phased arrays and seismic waves—would generalize the latent Earth observation space. Third, expanding its support for downstream tasks, such as investigating how the latent observation space drives cross-modal and cross-instrument tasks, would further enhance its utility. Lastly, improving interpretability by analyzing the relationships between different observations within the latent space could unveil underlying atmospheric phenomena, advancing scientific discovery.

The potential implications of \name~ for observation-driven Earth modeling are substantial. The latent observation space holds the potential to serve as the foundational data framework for next-generation atmospheric tasks, such as precipitation nowcasting and polar route planning. By retaining rich observational information while facilitating cross-observation complementarity, it breaks the spatiotemporal limits of raw observational data. Moreover, due to its broad support for diverse observation types, incorporating additional data becomes straightforward. This represents a significant step forward in making rapid multisource observation fusion accessible to anyone.




\newpage
\clearpage

\section{Method}
Compared with grid data, training the \name~ system for atmospheric observations is substantially more challenging due to the diversity of observing instruments and the heterogeneity of the resulting measurements. 
Atmospheric observation datasets vary widely in physical meaning, number of channels, spatiotemporal coverage, and data structures, unlike the uniform structure typically found in gridded data.
To accommodate heterogeneous, multi-source, and multi-modal Earth system observations, we design the \name~ system, which first tokenizes the heterogeneous data and then employs a flexible MMAE (Mutimodal Masked Auto Encoder)~\cite{mizrahi20234m} to learn a fused latent observational space. 
On top of this latent observational space, a Transformer is used for prediction, and an inversion module is introduced to support a variety of downstream tasks.

\subsection{Tokenization of heterogeneous observations}

\paragraph{Satellite observations tokenization}

The satellite data used in \name~ covers atmospheric satellite observations from both geostationary and polar-orbiting satellites. 
A swath of satellite observation data at a single specific acquisition time can be represented as $\overline{x}_{sat} \in \mathcal{R}^{C_{sat} \times H_{sat} \times W_{sat}}$. 
Due to the high resolution and wide coverage of satellite observations, $H_{sat}$ and $W_{sat}$ are relatively large. 
To efficiently extract small-scale information and richer tokens from satellite observations, tokenization is performed using a sliding window over global satellite sub-images of size $H_{sub}$ and $W_{sub}$ denoted as $x_{sub} \in \mathcal{R}^{C_{sat} \times H_{sub} \times W_{sub}}$. 

A further challenge in tokenizing atmospheric satellite observations is the incomplete and temporally variable coverage, particularly for polar-orbiting satellites.
Their orbital motion results in swatch-like observations that cover only (about 5\%) of the global atmosphere at any given time, with these swaths shifting continuously.
Therefore, we concatenate the missing values' mask, $m_{sub}$, as an additional feature to the satellite channels. 

Since the satellite observation channels are highly related to weather analysis products, we use a transformer encoder and decoder, specifically designed for weather data compression~\cite{han2025climate}, for tokenization of satellite observations.
Specifically,  $[x_{sub}, m_{sub}]$ is first projected by a $p \times p$ convolution to obtain the $embed_{sub}$. 
The resulting $embed_{sub}$ is then processed through a stack of $L$ stacked masked attention blocks, where the attention mask $m_{attn}$ is obtained by applying $p \times p$ area downsampling to $m_{sub}$.
The output is passed through a dimensionality-reduction projection layer, producing tokens $z_{sub} \in \mathcal{R}^{4C_{sat}\times \frac{H_{sat}}{p} \times \frac{W_{sat}}{p}} $.

To preferentially retain tokens in $z_{sub}$ that contain more valid information, we perform a morphological erosion on $m_{attn}$ for token selection. 
This operation not only removes tokens with no observations, but also further discards tokens located at the boundaries of satellite swaths, which only partially contain observed data. 
In \name, the satellite tokenizers are pre-trained separately in advance, with the objective function given by:
\begin{equation}
\mathcal{L}(\theta, \phi; x)
= \underbrace{\|(x_{sub} - \hat{x}_{sub}) \cdot m_{rec}\|_1^1}_{\text{reconstruction loss}}
+ \underbrace{\beta\mathrm{KL}\big(q_\phi(z_{sub}\mid x_{sub})\,\|\,p(z_{sub})\big)}_{\text{KL term}}. 
\label{eq:vae loss}
\end{equation}
In \Cref{eq:vae loss}, the first term is a reconstruction term: the decoder $S_{\theta}$ is encouraged to reconstruct the selected original observation $x_{sub}$ from tokens $z_{sub}$ that contain valid information.
Specifically, the filtered tokens are processed through $L$ mask attention blocks to obtain a reconstruction.
The loss is then computed between this output and the ground-truth $x_{sub}$ only over the valid region defined by the mask $m_{rec}$.
The second term is a KL-divergence(Kullback-Leibler divergence) regularizer that enhances the robustness of the satellite tokens, where $q_\phi(z_{sub}\mid x_{sub})$ is the token encoder, $p(z_{sub})$ is modeled as a Gaussian distribution and $\beta$ is set to $1 \times 10^{-6}$.

\paragraph{In-situ observations tokenization.}

\name~ also ingests in-situ observational data collected from a variety of platforms, including surface weather stations, marine vessels, and meteorological aircraft.
These observations at a single time step can be represented as a set of samples $\{(x_{i}^{insitu}, y_{i}^{insitu})\}$, where $x_{i}^{insitu} \in \mathcal{R}^{3}$ represents the spatial metadata of the observation (longitude, latitude, and altitude), and $y_{i}^{insitu} \in \mathcal{R}^{C}$ denotes the corresponding vector of physical variables, such as U-component wind speed, temperature, and relative humidity. 
In analogy to the satellite tokenizer, we also train an in-situ tokenizer on sliding subgraph windows of these observations.

\textbf{Encoding}: Since in-situ observations are represented as an unstructured set, we draw inspiration from bird’s-eye-view (BEV) representations~\cite{li2024bevformer} in point-cloud processing and aggregate these in-situ observations onto a set of spatial anchors before mapping them into in-situ tokens. 
We begin by generating grid anchors $x^{grid}$ within each sub-image. 
Specifically, we uniformly partition the longitude–latitude range covered by the sub-image into an $L \times L$ grid, yielding $x^{grid} \in \mathcal{R} ^{L \times L}$. 
The positional features of these anchors, denoted as $f_{i}^{grid}$, are obtained through a two-layer feed-forward network (FFN). 
We then randomly sample $N$ observations from $\{(x_{i}^{insitu}, y_{i}^{insitu})\}$ and aggregate them with the anchor positional features. 
For each sampled observation, missing values in $y_{i}^{insitu}$ are imputed, and the data is projected to a feature vector $f_{i}^{insitu}$. 
The altitude component in $x_{i}^{insitu}$ is subsequently used to modulate $f_{i}^{insitu}$ via a conditional layer normalization, thereby injecting vertical information into the feature representation.
We then compute the haversine distance between each observation's longitude and latitude and the anchor positions.
For each anchor, $K$ nearest observations are selected based on this distance metric using K-nearest-neighbor (KNN).
Finally, each anchor aggregates features from its $K$ nearest observations using a point-attention mechanism~\cite{zhao2021point}:
\begin{align}
\mathbf{f}_{grid_i} &= 
\sum_{x_j^{insitu} \in \mathcal{N}(grid_i)} 
\rho\!\left( 
    \gamma(\varphi(f_i^{grid}) - \psi(f_j^{insitu}) + \delta)
\right)
\odot 
(\alpha(f_j^{insitu}) + \delta), \\
\delta &= \theta(x_i^{grid} - x_{j}^{insitu}),
\end{align}
where $\varphi, \psi, \alpha$ denote point-wise feature transformations implemented as linear projections, $\delta$ is a positional encoding function, and $\rho$ is a normalization function, such as softmax. This point-attention module can be viewed as a set operator that acts on a collection of feature vectors and is therefore directly applicable to the entire set of in-situ observations. 
To spatially align with the coarser satellite tokens $z_{sat}$, the grid-level features $\mathbf{f}_{grid}$ are further downsampled by a stack of ResNet blocks with self-attention layers~\cite{rombach2022high}, producing the in-situ tokens \textbf{$z^{insitu}$}. 

\textbf{Decoding}: 
During the decoding stage, the in-situ tokens are first mapped to grid features $\hat{f}^{grid}$, facilitating continuous spatial decoding of physical variables at arbitrary query locations. 
Specifically, we stack several ViT (Vision Transformer)~\cite{dosovitskiy2020image} blocks with upsampling convolutions to transform the tokens into higher-resolution grid features.
To decode the physical variables at a specific location, we utilize a three-dimensional meta query vector $(x, y, z)$, where $x$ and $y$ denote the longitude and latitude of the site and are used to decode positional information from the grid features, while $z$ represents altitude or other auxiliary geographic attributes. 
The latter is injected via conditional layer normalization, allowing the model to explicitly capture vertical variations within the feature space. 

To facilitate off-grid predictions at any station locations, we adopt a setconv decoder~\cite{garnelo2018neural}.
Given the spatial coordinates of a target station, the decoder aggregates the corresponding grid features by geo-distance–based kernel weighting to obtain the location feature of that station. 
To maintain feature disentanglement and prevent cross-interference, different meta queries use independent parameterizations and sampling strategies.
Specifically, we estimate the probability density function of the target variables using a kernel regression formulation to sample the station-level features: 
\begin{equation}
    \hat{f}_{i}^{insitu} = \frac{\sum_n \hat{f}_{n}^{grid} \Psi( x_{i}^{insitu} - x_{n}^{grid})}{\sum_n \Psi(x_{i}^{insitu} - x_{n}^{grid})},
\end{equation}
where $\{(x_{n}^{grid}, y_{n}^{grid})\}$ denote the longitude–latitude coordinates of the grid cells and their associated features. This normalized kernel-weighting scheme fully exploits local neighborhood information while ensuring spatial smoothness and stability of the predictions. Finally, an MLP maps the sampled in-situ feature $\hat{f}_{i}^{insitu}$ to the physical variables $\hat{y}_{i}^{insitu}$ in the observation space.

\textbf{Joint Training}: 
To mitigate the effects of station sparsity on the station tokens, we adopt a joint mask training strategy with the multimodal masked autoencoder.
By performing tokenization jointly with MMAE and leveraging dense satellite observations to supplement the sparse in-situ stations, the model learns station tokens that are robust to the irregular spatial distribution of stations.

Specifically, for each pair $\{(x_{insitu},y_{insitu})\}$, we sample a random binary mask $M_{rand}$ and apply it to station observations with the same masking ratio as in satellite masks. 
The partially masked pairs $\{(x_{insitu},y_{insitu})\cdot M_{rand}\}$ are fed into the in-situ encoder $\mathcal{E}(\cdot)$, and the decoder $\mathcal{D}(\cdot)$ reconstructs the in-situ targets conditioned on the meta queries $\{x_{insitu}\}$:
\begin{equation} 
    \hat{y}_{insitu} = \mathcal{D}(\mathcal{E}(\{(x_{insitu}, y_{insitu})\cdot M_{rand}\}), \{x_{insitu}\}).
\end{equation}
This design explicitly forces the tokenization of in-situ observations to contain transformations from satellite observations to station observations. The joint in-situ loss is defined as the L\-1 reconstruction error between the ground-truth and predicted station variables:
\begin{equation}
     \mathcal{L}_{insitu\_joint} = \|(y_{insitu} - \hat{y}_{insitu}) \|_1^1,
\end{equation}
encouraging accurate recovery of masked station signals while effectively exploiting the auxiliary information provided by satellite measurements.

\subsection{MMAE modality fusion module}


\begin{figure}[ht]
    \centering
    \includegraphics[width=1.0\linewidth]{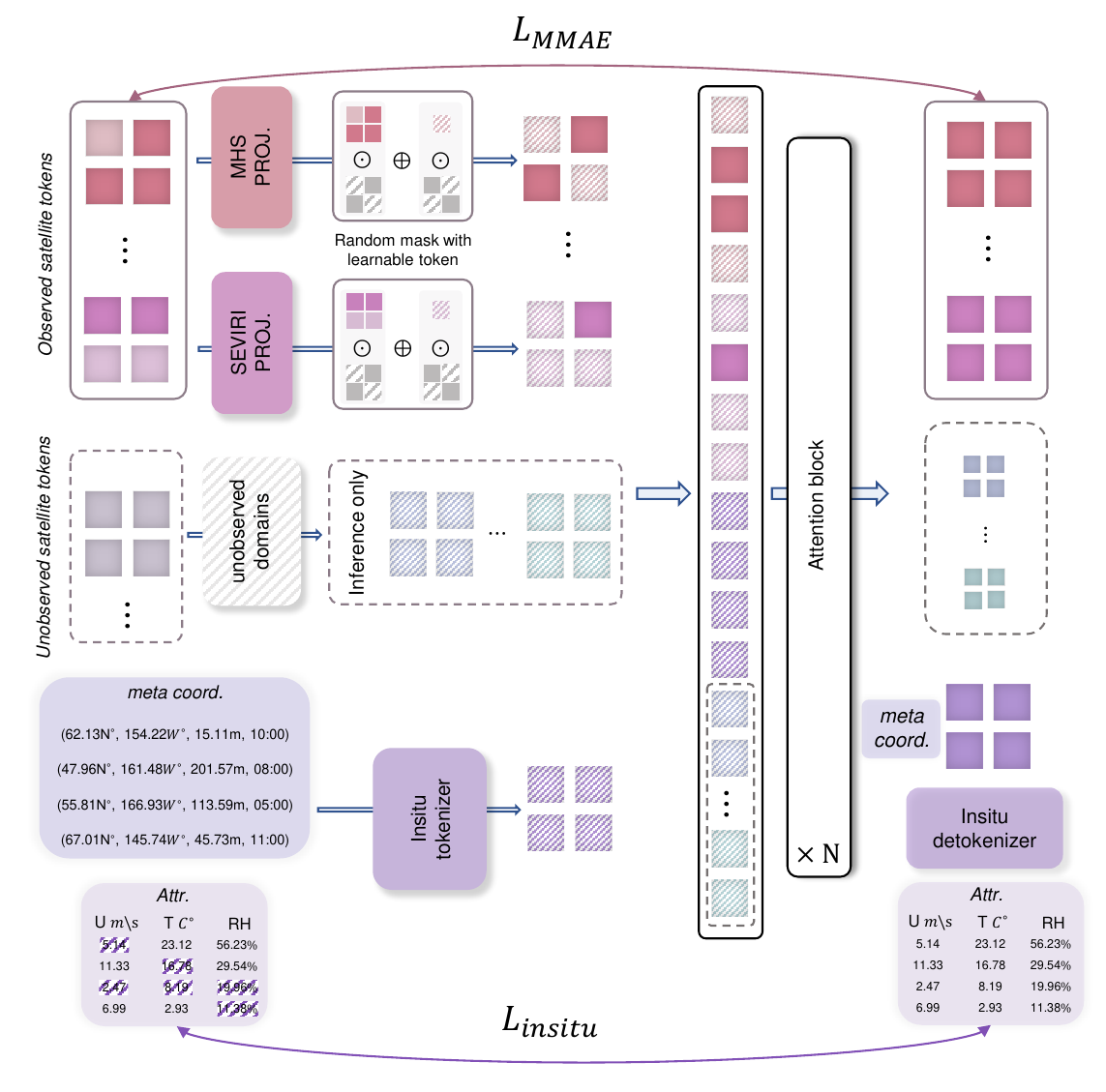}
    \vspace{-0.7cm}
    \caption{
    The structure of the MMAE modality fusion module.
    }
    \label{method:mmae}
\end{figure}

MMAE modality fusion module integrates observation tokens $z^m$ from multiple data sources into a unified latent observational space, where $m \in \{m_0, m_1, \cdots, m_{n}\}$ denotes the specific modality.
To enable fusion in a unified latent observational space, MMAE of \name~ first applies a set of modality-specific projectors $\{\mathcal{E}_{m_0}, \cdots, \mathcal{E}_{m_n}\}$ that map tokens from different instruments and modalities into the shared fusion space: 
\begin{equation}
    d_{m_i} = Norm_{m_i}(\mathcal{E}_{m_i}(z^{m_i})).
\end{equation}
In particular, to improve the stability of joint multi-modal training, we introduce a modality encoder norm, $Norm_{m_i}$, within each projector. 
This normalization mechanism constrains the encoded features at the modality level, making the outputs of different modalities more consistent in scale and distribution. 
As a result, it alleviates training instabilities caused by distributional discrepancies across observations and improves numerical stability and optimization convergence during modality fusion.

While the features from different modalities are projected into a unified latent observational space, the availability of observations from each instrument still exhibits substantial temporal and spatial variability.
To accommodate this, MMAE of \name~ supports a flexible configuration of input modalities:
At the sub-image level, arbitrary combinations of instrument tokens can participate in training, which maintains training efficiency while adapting to complex missing-observation patterns. 
Specifically, if a modality lacks valid tokens within the current temporal window of a sub-image, it is excluded from both the input and target sets.
In addition, we allocate a dedicated learnable domain token $\tilde{d}^{m_i}$ for each instrument and modality. 
These tokens are used to fill spatially sparse regions and positions masked by the mask modeling strategy, and during inference, they allow the fused representation to be propagated to instruments without observations, such as offline devices or uncovered regions of geostationary satellites. 

The overall process can be written as:
\begin{equation}
\{\hat{z}^{_0}, \cdots, \hat{z}^{m_n}\} = MMAE(\{(\mathcal{M}_{m_0} \cdot d^{m_0} + \sim\mathcal{M}_{m_0} \cdot\tilde{d}^{m_0}), \cdots, (\mathcal{M}_{m_n} \cdot d^{m_n} + \sim\mathcal{M}_{m_n} \cdot\tilde{d}^{m_n})\} ),
\end{equation}
where $M_{m_i}$ denotes the union of inherently missing observations of the $i-$th instrument and the additional masked positions generated by the mask modeling strategy. 
With this design, the model naturally supports arbitrary modality combinations at both training and inference time, and fully exploits the spatio-temporal complementarity of multi-source observations.

\textbf{Training}: MMAE trains and infers only one sliding window from the global range at a time, using data from a 12-hour time window to establish consistency in the observation representations across different time points.
In terms of the loss function design, MMAE employs a joint optimization objective that combines the reconstruction loss (MAE) with joint station training.
On the one hand, the MAE loss imposes reconstruction constraints on the masked tokens of each modality, defined as: 
\begin{equation}
    \mathcal{L}_{MMAE} =
\frac{1}{n} \sum_{j \leq n}
\frac{1}{|\mathcal{M}_{m_{j}}|}
\sum_{i \in \mathcal{M}_{m_{j}}}
\left\| \hat{z}_i^{m_{j}} - z_i^{m_{j}} \right\|_2^2,
\end{equation}
where $z_i^{m_{j}}$ and $\hat{z}_i^{m_{j}}$ represent the ground-truth and reconstructed tokens, respectively. 
On the other hand, we construct a joint station supervision loss, $\mathcal{L}_{insitu\_joint}$, based on in-situ observations to explicitly enforce the model's physical consistency and predictive accuracy at the observation station level. 
The final training loss for MMAE is defined as:
\begin{equation}
    \mathcal{L}_{MMAE} = \mathcal{L}_{MMAE} + \mathcal{L}_{insitu\_joint}.
\end{equation}
By simultaneously optimizing both the reconstruction task and the joint station supervision task, the model is able to maintain strong multi-modal representation capabilities while better capturing the relevant structures across modalities and spatial scales, thereby constructing a unified latent observational space for heterogeneous observations.

\subsection{Forecast module and inversion module}

To validate the atmospheric dynamics captured within the MMAE latent observational space and demonstrate their utility for downstream applications, \name~ includes a Forecast module and an Inversion module. The Forecast module performs temporal predictions within the latent observational space, while the Inversion module transforms these predictions into interpretable geophysical products, such as precipitation, carbon dioxide concentrations, and sea ice extent.

\textbf{Forecast}: During the forecasting phase, all observed tokens are input into MMAE without additional masks. 
The fused and completed representations generated by MMAE, denoted as $\{\tilde{z}^{m_0}, \cdots, \tilde{z}^{m_n} \}$, are used as the input data for the forecast module $\mathcal{F}$:
\begin{equation}
    [\hat{z}_{t+T:t+2T}^{m_{0}}, \cdots,  \hat{z}_{t+T:t+2T}^{m_{n}}] = \mathcal{F}([\tilde{z}_{t:t+T}^{m_{0}}, \cdots,  \tilde{z}_{t:t+T}^{m_{n}}]).
    \label{eq:forecast_module}
\end{equation}
As shown in \Cref{eq:forecast_module}, the Forecast module combines multi-modal observation representations from the previous $T$ time steps to predict the state for the subsequent $T$ time steps. 
Given that the atmosphere is a globally closed system, the predictions of the Forecast module are constrained on a global scale and not limited to a sub-image sliding window. 

The Forecast module consists of three main components: a unified patch embedding layer, attention blocks, and a domain-specific output decoder. 
First, the unified patch embedding layer concatenates the representations of all modalities along the channel and temporal window dimensions to form a unified token embedding. 
Then, inspired by~\cite{gong2025dawp}, the attention blocks leverage self-attention mechanisms to effectively learn the dynamics of atmospheric observation representations. 
Finally, a linear domain-specific output decoder is applied to each modality separately, transforming the unified token embedding into the corresponding modality tokens. 
The Forecast module computes the Mean Squared Error (MSE) across different modalities as the loss function:
\begin{equation}
    L_{pred} = \frac{\sum_{i\leq n}\|\tilde{z}_{t+T:t+2T}^{m_{i}} - \hat{z}_{t+T:t+2T}^{m_{i}}\|_2^2}{(n+1)\cdot h \cdot w},
\end{equation}
where $h$ and $w$ represent the number of tokens in the global latitude and longitude directions, respectively. 

\textbf{Inversion}: The Inversion module of \name~ inverts the predicted observations into products with L1 signals, across various fields such as sea ice, carbon dioxide concentrations, and precipitation. 
The inversion module is similar to the Satellite Observations Tokenization module, using a region sliding window strategy, but with the target replaced by downstream products.

\section{Dataset}

\textbf{In-situ observations} form the foundation of modern data assimilation for weather forecasting, as they provide direct measurements with relatively small errors. 
Despite the increasing availability of satellite data, these in-situ observations remain indispensable for constraining the atmospheric state.
Specifically, our framework integrates radiosonde soundings from the Integrated Global Radiosonde Archive (IGRA) and other operational sonde networks, surface reports from the METAR system, and aircraft observations mainly provided through the Aircraft Meteorological Data Reports (AMDAR) and the Aircraft Communications Addressing and Reporting System (ACARS). 
Together, these datasets (detailed in Supplementary Materials) ensure reliable coverage of temperature, humidity, pressure, and wind at critical synoptic locations.

\textbf{Polar-orbiting satellites} fly at low Earth orbits (LEO) and provide observations at finer spatial resolution than geostationary satellites, making them indispensable for global NWP systems. 
They are equipped with a wide range of instruments, and among these, the most important are microwave and infrared sounders such as the Advanced Microwave Sounding Unit-A (AMSU-A) and the Microwave Humidity Sounder (MHS).
These sensors deliver detailed vertical profiles of atmospheric temperature and humidity that have been proven essential for global assimilation. 
In addition, microwave and infrared imagers, for example the Special Sensor Microwave Imager/Sounder (SSMIS), provide information on precipitation, cloud liquid water, sea ice, surface winds, and land/sea surface properties. 
The Advanced Scatterometer (ASCAT) supplies high-quality ocean surface wind vectors, complementing the sounding and imaging data. 
To overcome the inherent revisit-time limitations of individual LEO platforms (typically two passes per day), we include multiple polar-orbiting satellites in our dataset  (summarized in Supplementary Materials) to ensure global spatio-temporal continuity.

\textbf{Geostationary satellites} complement this system by providing high-frequency, continuous monitoring from fixed longitudinal positions, a capability vital for capturing the lifecycle of rapidly evolving mesoscale systems. 
While offering a more constrained set of observables than their polar-orbiting counterparts, their high temporal resolution is significant for dynamic tracking. 
In this study, we leverage data from Himawari-8 and the GridSat archive to bridge the temporal gaps in the polar-orbiting record, ensuring a seamless and comprehensive observation.

\backmatter












\section*{Code availability}

The project page is accessible at \url{https://earth-o1.github.io/Earth-o1/}. Supplementary materials can be found at \url{https://github.com/jasong-ovo/earth-o1-paper/}.







\bibliography{sn-bibliography}

\end{document}